\documentclass[acmsmall, review=false]{acmart}

\usepackage{soul}
\usepackage{multirow}
\usepackage{booktabs}
\usepackage{graphicx}
\usepackage{subfigure} 
\usepackage{float}
\usepackage{xspace}
\usepackage{bm}

\usepackage{amsmath}
\usepackage{amsfonts}

\DeclareMathOperator*{\minimize}{minimize}

\newcommand{\system}{HCMS\xspace}
\newcommand*{\eg}{\emph{e.g.}\@\xspace}
\newcommand*{\etc}{\emph{etc}\@\xspace}
\newcommand*{\ie}{\emph{i.e.}\@\xspace}
\newcommand*{\vs}{\emph{v.s.}\@\xspace}

\newcommand*{\etal}{\emph{et al.}\@\xspace}
\newcommand{\fcvid}{{\scshape FCVID}\xspace}
\newcommand{\anet}{{\scshape ActivityNet}\xspace}

\makeatletter
\newcommand\figcaption{\def\@captype{figure}\caption}
\newcommand\tabcaption{\def\@captype{table}\caption}

\makeatother

\AtBeginDocument{%
  \providecommand\BibTeX{{%
    \normalfont B\kern-0.5em{\scshape i\kern-0.25em b}\kern-0.8em\TeX}}}

\setcopyright{acmcopyright}
\acmJournal{TOMM}
\acmPrice{15.00}\acmDOI{10.1145/3572776}
\begin{document}

\title{HCMS: Hierarchical and Conditional Modality Selection for \\ Efficient Video Recognition}

\author{Zejia Weng}
\affiliation{%
  \institution{Shanghai Key Lab of Intelligent Info. Processing, School of CS, Fudan University}
  \city{Shanghai}
  \country{China}}
\email{zjweng20@fudan.edu.cn}

\author{Zuxuan Wu$^{\dagger}$}
\affiliation{%
  \institution{Shanghai Key Lab of Intelligent Info. Processing, School of CS, Fudan University}
  \city{Shanghai}
  \country{China}}
\email{zxwu@fudan.edu.cn}

\author{Hengduo Li}
\affiliation{%
  \institution{Department of Computer Science, University of Maryland}
  \city{College Park, MD}
  \country{USA}}
\email{hdli@umd.edu}

\author{Jingjing Chen}
\affiliation{%
  \institution{Shanghai Key Lab of Intelligent Info. Processing, School of CS, Fudan University}
  \city{Shanghai}
  \country{China}}
\email{chenjingjing@fudan.edu.cn}

\author{Yu-Gang Jiang}
\affiliation{%
  \institution{Shanghai Key Lab of Intelligent Info. Processing, School of CS, Fudan University}
  \city{Shanghai}
  \country{China}}
\email{ygj@fudan.edu.cn}

\renewcommand{\shortauthors}{Zejia Weng and Zuxuan Wu, et al.}

\begin{abstract}
    Videos are multimodal in nature. Conventional video recognition pipelines typically fuse multimodal features for improved performance. However, this is not only computationally expensive but also neglects the fact that different videos rely on different modalities for predictions. This paper introduces Hierarchical and Conditional Modality Selection (HCMS), a simple yet efficient multimodal learning framework for efficient video recognition. HCMS operates on a low-cost modality, \ie, audio clues, by default, and dynamically decides on-the-fly whether to use computationally-expensive modalities, including appearance and motion clues, on a per-input basis. This is achieved by the collaboration of three LSTMs that are organized in a hierarchical manner. In particular, LSTMs that operate on high-cost modalities contain a gating module, which takes as inputs lower-level features and historical information to adaptively determine whether to activate its corresponding modality; otherwise it simply reuses historical information. We conduct extensive experiments on two large-scale video benchmarks, FCVID and ActivityNet, and the results demonstrate the proposed approach can effectively explore multimodal information for improved classification performance while requiring much less computation.
\end{abstract}



\maketitle

\section{Introduction}

The popularity of mobile devices and social platforms has spurred an ever-increasing number of Internet videos: more than $720,000$ hours of videos were uploaded to YouTube every day in 2019. The sheer amount of videos demands automated approaches that can analyze video content effectively and efficiently for a wide range of applications like web-video search,  video recommendation, marketing, \etc. However, video categorization is a challenging problem since compared to static images, videos are multimodal in nature and thus convey more information, including appearance information of objects in individual frames, motion clues among adjacent frames that depict how objects move, and acoustic information that is complementary to visual clues. 

Such a multimodal nature in videos has motivated extensive work to leverage inputs of different modalities for better video understanding. Simonyan \etal feed optical flow images to 2D convolutional neural networks (CNNs) for spatio-temporal feature learning~\cite{DBLP:conf/nips/SimonyanZ14}, and the explicit modeling of motion information has inspired numerous approaches~\cite{Feichtenhofer16,icmr15:eval2stream,lin2019tsm,wang2017non, zhang2016real, yang2020sta, song2020modality,ullah2018activity}. Unlike these approaches that take optical flow images as inputs, another line of work focuses on learning motion features directly from RGB images by modeling temporal relationships across frames with 3D convolutional layers~\cite{C3D,tran2018closer,xie2018rethinking,feichtenhofer2019slowfast,feichtenhofer2020x3d,wang2021multi}. Apart from visual information, audio signals are also utilized to improve classification accuracy~\cite{MVA:audiovisual,wu2016multi,gao2020listen,korbar2019scsampler} for video recognition, as well as tasks like self-supervised learning~\cite{arandjelovic2017look,owens2018audio}.

\begin{figure}[tbp]
\centering
\includegraphics[width=0.7\linewidth]{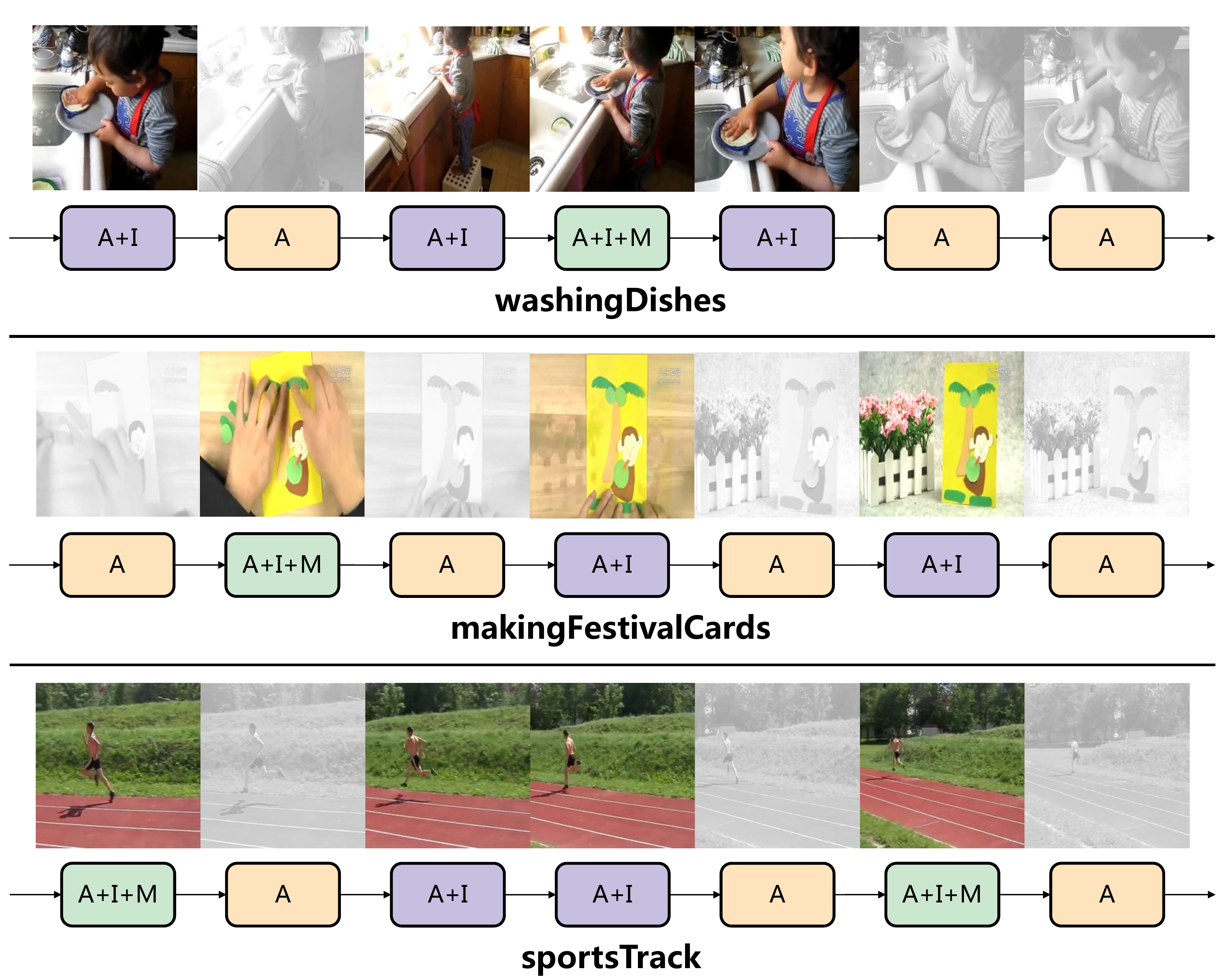}
\caption{A  conceptual overview of our approach. \system operates on the low-cost audio modality by default, and at each time step, it learns dynamically whether to use high-cost modalities on a per-input basis. Here, A: the audio modality; I: the appearance modality; M: the motion modality)}
\label{fig:teaser}
\end{figure}

How to effectively combine multimodal features remains a challenging problem for video classification. Existing work mainly performs feature fusion at two stages: (1) early fusion that simply concatenates different features as inputs to a neural network to generate predictions; (2) late fusion typically trains a separate model for each modality and then averages prediction scores from all models linearly or fuses the information from different modalities with adaptive weights generated by neural components. While using multimodal inputs generally improves recognition performance over simply using a single modality, it does not come for free---the computational cost grows dramatically, particularly for the motion modality. For example, when modeling motion information explicitly with optical flow images, we need to store and compute them beforehand. If motion clues are modeled with 3D convolutional networks, which operate on a stack of frames, the computational cost then increases by an order of magnitude. Furthermore, fusion approaches mentioned above typically combine multimodal features for all video clips. While an ideal multimodal fusion approach is expected to learn to assign a set of optimal weights for different modalities for improved performance, this is a difficult task as the only available supervision is video-level labels and there is no golden ground-truth indicating whether a particular modality is good or not. In addition, current fusion approaches produce one-size-fits-all weights for all videos in the dataset, regardless of the varying difficulties to make predictions for different videos.

In light of this, we ask the following questions: Is it possible to leverage the rich multimodal information without incurring a sizable computational burden? Do all videos require information from multiple modalities to make correct predictions? After all, different video categories require different \emph{salient} modalities to make correct predictions. For example, static appearance information suffice correct predictions for object-centric or scene-centric actions like ``playing billards'' and ``playing football'', while fine-grained motion information is needed to distinguish motion-intensive actions like ``long jump'' and ``triple jump''; for actions with unique audio cues like ``playing the piano'' and ``playing the flute'', visual information might not be needed at all! In addition, we postulate that modality needed differ even for samples within the same category due to large intra-class variations.  For instance, for the ``birthday'' event, some videos might contain salient audio clues like people singing a birthday song. As a result, the audio modality is critical for such videos while visual information is of great importance for birthday videos without the birthday song. This motivates us to learn modality fusion on a per-sample basis---deriving a modality usage policy to select different modalities conditioned on input videos not only to leverage the rich multimodal clues in videos for accurate predictions but also to save computation for improved efficiency (See Figure~\ref{fig:teaser} for a conceptual overview. For instance, the first video within Figure~\ref{fig:teaser} depicts the event of a child washing dishes. Since there are spatial redundancies in the video, \eg, the penultimate three consecutive images are very similar, our approach automatically ignores the computation of redundant visual information by skipping the last two frames, \ie, simply using audio information is sufficient.).

To this end, we introduce \system, a dynamic computation framework that adaptively determines whether to use computationally expensive modalities on a per-input basis for efficient video recognition. \system contains an audio, an appearance and a motion LSTM that are organized in a hierarchical manner, operating on the corresponding modality for temporal modeling. At each time step, the audio LSTM takes as inputs low-cost audio clues extracted by an audio net for a coarse evaluation of video content. To save computation, the appearance and the motion LSTMs are designed to be updated as infrequently as possible with a gating design. In particular, conditioned on historical information and the current audio frame, the appearance LSTM determines whether to compute visual information with a gating function. If visual information is needed, the appearance network will be activated and the corresponding LSTM will be updated; otherwise, \system goes to the next time step. A similar strategy is adopted to learn whether to compute motion information and update the motion LSTM. We then fuse hidden states from three LSTMs such that hidden states from the motion LSTM contain information seen so far and can be readily used for classification. Built upon the Gumbel-softmax trick, \system learns binary modality selection decisions in a differentiable manner.

We conduct extensive experiments on two large-scale video recognition datasets, \anet and \fcvid. We show that \system offers comparable and slightly better performance compared to existing multimodal fusion approaches while requiring only 56\% and 86\% computation on \fcvid and \anet, respectively. We also show qualitatively and quantitatively that videos samples have different preferences towards modalities.

\begin{figure*}[tbp]
\centering
\includegraphics[width=0.95\linewidth]{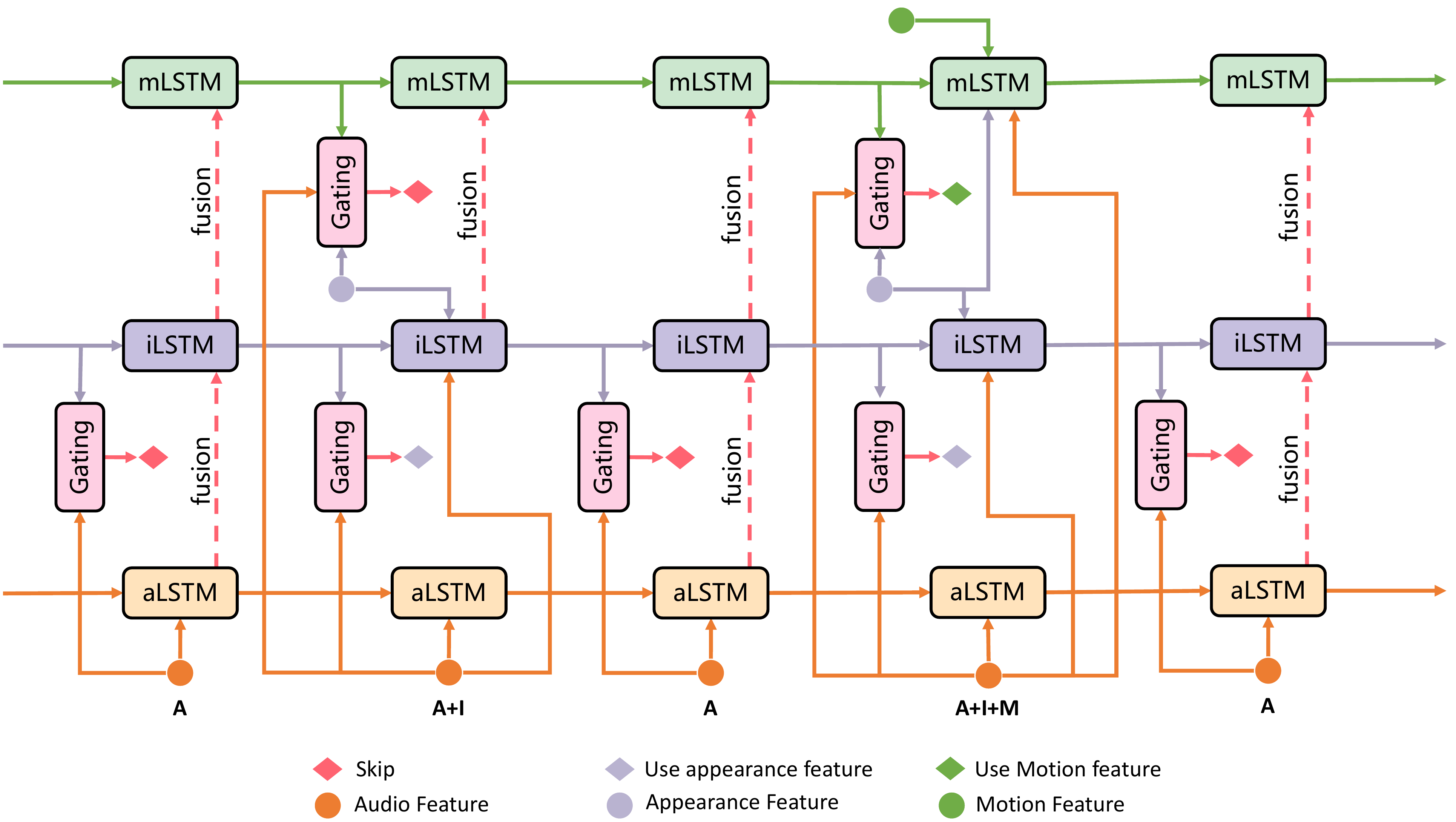}
\caption{An overview of our framework. At each time step, \system uses audio clues and historical information to determine whether to use the appearance modality. Similarly, the activation of the motion modality will be decided based on appearance features and historical clues. See texts for more details.}
\label{fig:framework}
\end{figure*}

\section{Related Work}

\vspace{1mm}
\noindent\textbf{Deep Networks for Video Recognition.} Deep neural networks have achieved great success on a wide range of tasks~\cite{he2016deep, weng2021semi, gao2020exploring, zhang2022local, zheng2022clustering, aloufi2022mmsum, carion2020end}. In the field of video recognition,  deep neural networks have also been extensively used over the past few years~\cite{DBLP:conf/nips/SimonyanZ14,C3D,wang2018temporal,wang2017non,tran2018closer,feichtenhofer2019slowfast,lin2019tsm,feichtenhofer2020x3d, xu2019dense,wang2021multi}. In particular, there is a plethora of work focusing on extending 2D networks for temporal modelling through either designing temporal aggregation methods applied to per-frame 2D spatial features, or inserting 3D convolutional layers to 2D networks to learn spatio-temporal features jointly. The former line of research aggregates spatial features along the temporal axis with pooling operations~\cite{DBLP:conf/nips/SimonyanZ14,wang2018temporal,Feichtenhofer16}, recurrent networks like long short-term memory (LSTM) network~\cite{ng2015beyond,li2018videolstm} or carefully designed modules like Temporal Relational Network (TRN)~\cite{zhou2018temporal} and Temporal Shift Module (TSM)~\cite{lin2019tsm}. 3D networks, on the other hand, take as inputs a snippet of frames (clips) and model temporal relationships across frames directly by a stack of 3D convolutional layers~\cite{C3D,tran2018closer,feichtenhofer2019slowfast,feichtenhofer2020x3d,wang2017non,carreira2017quo,wang2021multi}. While being conceptually simple and providing rich motion clues, 3D networks often demand more computation than 2D networks by at least an order of magnitude, due to the model complexity and the large number of snippets of frames needed to cover the entire video. In this paper, we use modalities that come at a lower computational cost and learn whether to use motion information conditioned on the current inputs and historical information.

\vspace{1.5mm}
\noindent\textbf{Efficient Networks for Video Recognition.}
There is a growing interest in designing efficient network architectures for both 2D and 3D networks in video recognition tasks. For 2D networks, common design choices of efficient architectures for image classification such as channel-wise separable convolutions in MobileNets~\cite{howard2017mobilenets,sandler2018mobilenetv2} and Shufflenet~\cite{zhang2018shufflenet} have been naturally adapted for video recognition. Notably, several light-weight temporal aggregation modules on top of 2D networks were introduced recently and achieve superior recognition performance, such as the temporal relational module in~\cite{zhou2018temporal} and the temporal shift module in~\cite{lin2019tsm}. For 3D networks, various approaches have been proposed with the main focus on inserting 3D convolutional layers wisely, \eg using them only for layers closer~\cite{xie2018rethinking,feichtenhofer2019slowfast} or further~\cite{tran2018closer,xie2018rethinking} to final layer and using a bottleneck structure~\cite{tran2018closer} to save computaion.~\cite{feichtenhofer2020x3d} also proposes to expand a tiny 2D network in multiple dimensions to produce a series of models with better accuracy/efficiency trade-offs, while \cite{jiang2015super} proposes to apply alternative efficient features and reduce redundant frames for faster video recognition.

Our work shares the same purpose with these existing work but approaches it in a different way. These approaches improve the network efficiency generally but use fixed parameters and the same amount of computation for all input videos, regardless of their large intra-class variations. On the contrary, our method is agnostic to network architectures and learns to generate on-the-fly decisions on whether to use computationally expensive modalities such that the amount of computation is reduced while the prediction remains correct. 

\vspace{1.5mm}
\noindent\textbf{Multimodal Video Understanding.} Apart from RGB frames, other modalities such as audio information and motion features are also leveraged in many existing approaches to improve video recognition performance. Two-stream ConvNets~\cite{DBLP:conf/nips/SimonyanZ14} use pre-computed optical flow images as extra inputs to deep networks, which then becomes a common paradigm for video recognition~\cite{wang2018temporal,Feichtenhofer16,lin2019tsm,zhu2018fine,jiang2018modeling,yang2016multilayer} especially in approaches using 2D networks. Audio cue is also utilized to improve video recognition performance~\cite{wu2016multi,xiao2020audiovisual,korbar2019scsampler}. These multimodal inputs are used in other video tasks such as self-supervised learning in videos~\cite{arandjelovic2017look,owens2018audio}, sound localization~\cite{tian2018audio,senocak2019learning} and generation~\cite{zhou2018visual,zhou2019vision} from videos as well. Existing fusion approaches typically use fixed fusion weights for all samples. Besides, sample-aware dynamic weight fusion approaches have also been explored in several studies. For instance, in the development of query-based vision retrieval system, researchers realize that multiple modalities have different importance for a particular query. Various kinds of query-aware dynamic fusion approaches are designed {\cite{wei2011concept, zheng2015query, hou2021conquer}} in which the effectiveness of different types of features is judged based on the query to produce the optimal fusion weights for each specific query. However, these methods focus on studying how to dynamically weight multimodal features, e.g. by giving small weights to unimportant features, instead of discarding them. In contrast, our method dynamically chooses whether to use higher-cost modality features for computation on a per-input basis, such that the rich information from multiple modalities are leveraged with much lower computational cost.

\vspace{1.5mm}
\noindent\textbf{Conditional Computation.} 
Conditional computation methods have been proposed for both image and video recognition to reduce computation while maintaining accuracy by dynamically performing early exiting in deep networks~\cite{bolukbasi2017adaptive,li2019improved}, varying input resolutions~\cite{yang2020resolution,uzkent2020learning}, skipping network layers or blocks~\cite{wang2017skipnet, zhu2020faster} and selecting smaller salient regions~\cite{uzkent2020learning}, \etc. An additional perspective introduced in video is to dynamically attend to informative temporal regions of the entire video, in forms of salient frames and clips as in AdaFrame~\cite{wu2019adaframe} and SCSampler~\cite{korbar2019scsampler}, to speed up inference. Audio cues are used in~\cite{gao2020listen} to ``preview'' video clips efficiently. While we also leverage audio information, our goal is determine whether use computationally expensive modalities at each time step. In addition, our approach is well-suited for both online and offline predictions where~\cite{korbar2019scsampler,gao2020listen} can only be used for offline recognition.
Some recent approaches~\cite{meng2021adafuse,pan2021va,liu2021selective} also explore dynamic temporal feature aggregation within 3D networks by fusing features along temporal axis controlled by gating functions conditioned on input clips, resulting in clear computational savings.

Closely related but orthogonal to the approaches above, our method improves the inference efficiency by selectively using features (instead of using all) from modalities that require larger computational resources to compute on a per-input basis in a multimodal input scenario. For inputs where low-cost features suffice for correct predictions, high-cost ones are skipped for a reduced computational cost. Such a dynamic selection for multimodal features is by design orthogonal to existing dynamic methods on selecting resolution and input regions, network layers, feature downsampling, \etc;  we believe our work is complementary to these approaches.

\section{Method}
Given a test video, our goal is to learn what modality to use at each time step conditioned on the inputs, such that we can fully exploit the multimodal clues in videos and at the same time save computation by using computationally expensive modalities as infrequently as possible.  We achieve this with a hierarchical and conditional modality selection framework, \system, which operates on three different modalities, including audio information, appearance clues in 2D images and motion information among adjacent frames. By default, \system uses audio features for fast inference, and learns whether to use appearance and motion features in a hierarchical manner.  Figure~\ref{fig:framework} presents an overview of the framework. Below, we first introduce three types of modalities used by \system, and elaborate how to select different modalities at each time step for effective and efficient predictions. 

\subsection{Preliminaries}

Given a training set with $N$ video clips that belong to $C$ classes, where each video is associated with three types of modalities, \ie, audio, appearance, and motion. Thus, we can represent a video with $T$ video snippets as:

\begin{align}
  \bm{V} = & [\bm{V}_1, \bm{V}_2, \bm{V}_t, \cdots, \bm{V}_T] \nonumber \\
  \text{where} \, & \bm{V}_t =   \{\bm{V}^{A}_t, \bm{V}^{I}_t, \bm{V}^{M}_t\}.
\end{align}

Here ${V}^{A}_t$, ${V}^{I}_t$, and ${V}^{M}_t$ denote the audio, appearance and motion modality, respectively.  At each time step, we extract features from these three modalities with three CNN models.

\vspace{1.5mm}
\textbf{Audio Network.} Audio signals often contain important video context information and is helpful to improve the accuracy of video classification. More importantly, audio information is generally computationally cheap to compute. To extract audio information for the $t$-th time step (we split the entire video into $T$ segments), we first resample the input audio to 16 kHZ, then use short-time Fourier transform to convert the 1-dimensional audio signal into a 2-dimensional spectrogram  with the size of $96\times 64$, and finally use a MobileNetV1~\cite{howard2017mobilenets} network pretrained on AudioSet-YouTube~\cite{gemmeke2017audio}, to transform the 2D spectrogram into a compact feature representation. Formally, we use $f_{\theta_A}: {V}^{A}_t \mapsto \bm{x}^{A}_t$ to map an audio spectrum to a feature vector, where $f_{\theta_A}$ denotes the weights of the audio network and $\bm{x}^{A}_t$ is the audio features produced by the network.

\vspace{1.5mm}
\textbf{Appearance Network.} We use a 2D CNN model to capture appearance information from single RGB frames. The intuition is that for videos that contain relatively static objects and scenes like ``panda'', ``giraffe'' and ``river'', appearance information modeled by 2D CNNs are generally sufficient. Compared to audio clues which are usually noisy, the appearance network provides decent visual information with a moderate computational cost and it suffices in most cases. To this end, we use an EfficientNet~\cite{tan2019efficientnet} backbone to extract appearance clues. Formally, we define the mapping from an RGB frame to a feature vector as $f_{\theta_I}: {V}^{I}_t \mapsto \bm{x}^{I}_t$, where $f_{\theta_I}$ is the weights of the appearance network and $\bm{x}^{I}_t$ is the appearance feature.

\vspace{1.5mm}
\textbf{Motion Network.} We adopt a 3D CNN model to capture motion clues that depict how objects move among stacked frames. Modeling motion information is important for understanding the differences between subtle actions like ``standing up'' and ``sitting down''. Operating on a stack of frames, the motion network $f_{\theta_M}: {V}^{M}_t \mapsto \bm{x}^{M}_t$ parameterized by weights $\theta_M$ produces the motion feature $\bm{x}^{M}_t$ with a sequence of 3D convolutions and temporal pooling. We instantiate the motion network with a SlowFast network~\cite{feichtenhofer2019slowfast}. It is worth pointing out that while the motion network is powerful for video recognition, it is extremely computationally expensive.

Conventional fusion approaches typically compute three types of features from different modalities at each time step. And then they combine them with feature concatenation to form a long vector as inputs to a multi-layer perceptron for classification. However, this is not only computationally demanding but also neglects the fact that different videos have preferences towards different modalities.

\subsection{Hierarchical and Conditional Modality Selection} \label{subsec:framework}
Recall that we aim to explore the multimodal information in videos for improved recognition 
in a computationally efficient manner such that computational resources are dynamically allocated conditioned on input samples. However, learning what modalities to use for each video sample at each time step is extremely challenging since (1) there is no direct supervision indicating whether a modality is necessary; (2) determining whether to use a particular modality involves making binary decisions, which is not differentiable and thus poses challenges for optimization; (3) this creates varying inputs for different samples over time---the number of modalities to be used is different, which is not GPU-friendly.

To mitigate these issues, we introduce a dynamic computation framework, which contains three LSTMs, an audio LSTM, an appearance LSTM and a motion LSTM, which work collaboratively for modality selection. Observing that the computation required to extract audio, appearance and motion information increases as the size of networks grows, these three LSTMs are organized in a hierarchical manner. Since audio clues are computationally efficient, \system uses audio information by default for an economical understanding of video contents. At each time step, based on historical information and the current audio frame, \system dynamically decides whether to compute appearance information for an examination of visual information. Then the appearance LSTM proceeds using a similar strategy---if appearance information is sufficient for recognizing the class of interest, \system simply goes to the next time step; otherwise, \system requires more detailed understanding of visual information which thus activates the motion network. Below, we introduce the three LSTMs and the modality selection module in detail.

\vspace{1.5mm}
\noindent \textbf{Audio LSTM} The audio LSTM takes in audio features to quickly analyze potential semantics in videos for a general overview. Formally, given $\bm{x}^{A}_t$ of the $t$-th time step extracted by the audio network, hidden states ${\bm h}^A_{t-1}$ and cell states ${\bm c}^A_{t-1}$ from the previous time step , the audio LSTM computes the hidden and cell states of the current time step as:
\begin{equation}
\label{eq:lstm_a}
{\bm h}_t^A, \, {\bm c}_t^A = \texttt{aLSTM}({\bm x}^A_t, \, {\bm h}_{t-1}^A, \, {\bm c}_{t-1}^A).
\end{equation}

\vspace{1.5mm}
\noindent \textbf{Appearance LSTM}. Unlike standard LSTMs, the appearance LSTM contains a gating module which decides whether to compute appearance features conditioned on current audio inputs, and historical clues. In particular, the gating module is parameterized by weights $\bm{W}_I$ and generates a binary decision:
\begin{align}
  \bm{g}^I_t & = {\bm{W}_I}^T[{\bm x}^A_t; {\bm h}_{t-1}^I ; {\bm c}_{t-1}^I] \in \mathbb{R}^2,
\end{align}
where $[;]$ denotes the concatenation of features.

Since the gating decisions are discrete, one can select the entry with a higher value, however this is not differentiable. Instead, we define a random variable $G^I_t$ to make decisions through sampling from $\bm{g}^I_t$. When $G^I_t=0$, the appearance LSTM simply reuses previous information and the computation of visual features is skipped. When $G^I_t=1$, the appearance network is used to  extract visual information; the appearance LSTM then takes concatenated audio and appearance features together with hidden and cell states from a previous time step, to generate hidden and cell states for the current time step. More formally, the updating strategy of the appearance LSTM can be written as:
\begin{equation}
\label{eq:lstm_i}
{\bm h}_t^I, \, {\bm c}_t^I = 
\begin{cases}
 \texttt{iLSTM}([{\bm x}^A_t;  {\bm x}^I_t], \, {\bm h}_{t-1}^I, \, {\bm c}_{t-1}^I), \,\, \text{if} \,\, G^I_t=1 \\
 {\bm h}_{t-1}^I, {\bm c}_{t-1}^I, \,\, \text{otherwise}
\end{cases}
\end{equation}

\vspace{1.5mm}
\noindent \textbf{Motion LSTM}. The motion LSTM contains a similar gating module as the appearance LSTM to decide whether to capture motion information for the current step. In particular, the gating module is parameterized by weights $\bm{W}_M$ and generates a binary decision:
 \begin{align}
  {\bm g}^M & = {\bm{W}_M}^T[{\bm x}^A_t; {\bm x}^I_t; {\bm h}_{t-1}^M ; {\bm c}_{t-1}^M] \in \mathbb{R}^2.
\end{align}

A random variable is similary defined $G^M_t$ to sample from $\bm{g}^M_t$ for decision making. When more information (\ie, $G^M_t=1$) is required from the current step, the motion network will be used; otherwise \system proceeds to the next time step.
\begin{equation}
\label{eq:lstm_m}
{\bm h}_t^M, \, {\bm c}_t^M = 
\begin{cases}
 \texttt{mLSTM}([{\bm x}^A_t; {\bm x}^I_t;  {\bm x}^M_t], \, {\bm h}_{t-1}^M, \, {\bm c}_{t-1}^M), \,\, \text{if} \,\, G^M_t=1 \\
  {\bm h}_{t-1}^M, {\bm c}_{t-1}^M, \,\, \text{otherwise}.
\end{cases}
\end{equation}

\vspace{1.5mm}

\noindent \textbf{Fusing hidden states}. The motion LSTM contains the most powerful features, and thus we use the hidden states from the motion LSTM to make classification predictions. However, since we wish to save computation, the motion LSTM is expected to make updates only when necessary. As a result, information in the motion LSTM is not complete, since for some steps it is not activated. Instead, such information is captured by the audio LSTM and the appearance LSTM.  To remedy this, we perform a recursive fusion of hidden states. In particular, when $G^I_t=0$, we simply copy the hidden states of the audio LSTM to override a part of hidden states in the appearance LSTM. Similarly, when $G^M_t=0$, we copy the hidden states from the appearance LSTM to the motion LSTM. As a result, the hidden states from the motion LSTM contain information seen so far by all LSTMs, and can be readily used for classification.

\subsection{Objective Function}
As our goal is to fully leverage the rich multimodal information while using computation-demanding modalities as infrequently as possible, we devise the loss function to incentivize correct predictions with less usage of high-cost modalities at the same time. Formally, the hierarchical LSTM networks consist of learnable parameters $\bm{W} = \{\bm{\Theta}_A, \bm{\Theta}_I, \bm{\Theta}_M, \bm{W}_I, \bm{W}_M\}$, where $\bm{\Theta}_A, \bm{\Theta}_I, \bm{\Theta}_M$ denote parameters of audio, appearance and motion LSTMs and $\bm{W}_I, \bm{W}_M$ denote parameters of two gating modules. Given video $\bm{V}$ and its corresponding label $\bm{y}$, predictions $\bm{p}_T$ at the final time step $T$ is obtained and a standard cross-entropy loss is computed:

\begin{align}
    \bm{L}_{ce} = -\bm{y}\,\texttt{log}(\bm{p}_T(\bm{V}; \bm{W}))
\end{align}

The final predictions are produced with modality usage strategies $\sum_{t=1}^{T} G_t^I$ and $\sum_{t=1}^{T} G_t^M$ from gating modules applied at each time step. While $G_t$ can be readily sampled from a Bernoulli distribution given the probability generated by gating modules during training, \ie $G_t\sim\texttt{Bernoulli}(\bm{g}_t;\bm{W}_g)$, optimization of learning binary decisions is challenging as it is non-differentiable. It is feasible to formulate it as a reinforcement learning problem and train the network with policy gradient methods~\cite{sutton1998reinforcement}, yet the optimization requires carefully and manually designed rewards and is known to suffer slow convergence and large variance~\cite{sutton1998reinforcement,jang2016categorical}. 

To this end, we alternatively adopt Gumbel-Softmax trick~\cite{maddison2016concrete} to relax the non-differentiable sampling process from discrete distributions into a differentiable one. Specifically, at time step $t$, the discrete decisions derived as follows:

\begin{align}
GB_t = [GB_{t,1}, GB_{t,2}, ...,& GB_{t,K}] \\ 
{G}_{t,k} = \frac{\texttt{exp}((\texttt{log}\,g_{t,k}+ GB_{t,k})/\tau)} {\sum_{j=1}^{N} \texttt{exp}((\texttt{log}\,g_t^j + GB_t^j) / \tau)} & \quad \textnormal{for } k = 1,..,K
\label{eqn:gumbelsoftmax}
\end{align}

where $K$ denotes the number of categories and $K=2$ in our case of binary decision, and $GB_i =  -\texttt{log}\,(-\texttt{log}\,(U_i))$ denotes the gumbel noise with $U_i$ sampled form i.i.d distribution $\texttt{Uniform}\,(0, 1)$. To control the sharpness of the generated distribution vector $G_t$~\footnote{we omit the superscript $I$ and $M$ for brevity}, a temperature parameter $\tau$ is applied.

To save computation, we additionally introduce an additional loss that encourages a reduced usage of computationally expensive modalities:

\begin{align}
    \bm{L}_{usage} = (\frac{1}{T}\sum_{t=1}^{T} G_t^I - \gamma_1)^2 + (\frac{1}{T}\sum_{t=1}^{T} G_t^M - \gamma_2)^2
\label{eqn:loss_usage}
\end{align}

where $\gamma_1$ and  $\gamma_2 \in [0, 1]$  indicate a targeted budget of using appearance and motion networks, respectively. Finally, the overall objective is to minimize a weighted sum of both loss functions controlled by $\lambda$:
\begin{align}
    \minimize_{\bm{W}} \bm{L} = \bm{L}_{ce} + \lambda \ \bm{L}_{usage}
\label{eqn:loss_all}
\end{align}

In summary, at each time step $t$, we use outputs from Gumbel-Softmax twice Eqn.~\ref{eqn:gumbelsoftmax} to derive gating decisions. This is to connect the audio, appearance and the motion LSTM and decide whether to use computationally expensive modalities. Finally, at time $T$, the classification predictions are obtained along with the accumulated gating decisions to minimize the loss function defined in Eqn.~\ref{eqn:loss_all}.

\section{Experiments}
\subsection{Experimental Setup}
\subsubsection{Datasets}

To benchmark the effectiveness of our method, we use two challenging datasets namely \anet~\cite{caba2015activitynet} and \fcvid (Fudan-Columbia Video Dataset)~\cite{TPAMI-fcvid}. For \anet, we use the v$1.3$ split which contains $10,024$ training videos, $4,096$ validation videos and $5,044$ testing videos covering $200$ action classes. Since labels for testing videos are not publicly available, we report results on the validation set. \fcvid contains $45,611$ training videos and $45,612$ testing videos that belongs to $239$ categories. Videos in \anet and \fcvid are untrimmed, lasting $117$ and $167$ seconds on average respectively.

\subsubsection{Evaluation} For evaluation, we compute average precision (AP) for each category, and then average them across all categories to produce mean average precision (mAP) as instructed in~\cite{caba2015activitynet,TPAMI-fcvid}. Following common practices~\cite{wang2017non,feichtenhofer2019slowfast,wu2019adaframe}, we uniformly sample $T = 10$ or $128$ time steps (views) from each video.

\subsubsection{Implementation details}

We extract Audio features with MobileNetV1~\cite{yamnet_repo} which is pretrained on AudioSet-YouTube corpus. The appearance and motion networks are EfficientNet-B3~\cite{tan2019efficientnet} and SlowFast~\cite{feichtenhofer2019slowfast}  which are pretrained on ImageNet and Kinetics respectively. Then they are further finetuned on \fcvid and \anet. The input size to MobileNetV1, EfficientNet and SlowFast models are $96 \times 64$, $224 \times 224$ and $8 \times 224 \times 224$ ($8$ denotes number of frames in each clip), requiring  $0.07$, $0.99$ and $65.7$ GFLOPS respectively for feature extraction at each time step. In our framework, we first use a fully connected layer to project the $1024$-d audio feature, the $1536$-d appearance feature and the $2304$-d motion feature into the dimensions of $512$, $1024$ and $2048$. The features are then fed into the hierarchical LSTMs with $128$, $512$ and $2048$ hidden units. We implement \system with PyTorch and use Adam as the optimizer. The initial learning rate of the training schedule is set to 1e-4, which is multiplied by $0.92$ after every epoch. we set the controlling factors (in Eqn.~\ref{eqn:loss_usage}) $\gamma_1 \in [0.6, 0.85]$ and $\gamma_2 \in [0.94, 0.98]$. $\lambda$ in Eqn.~\ref{eqn:loss_all} is set to 2. In both training and testing phases, we extract 128 audio segments for each video and ensure the feature sequence can cover the entire video sequence. The corresponding 128 time steps will be the potential locations to calculate image and motion features.

\begin{table*}[t!]
\centering

  \caption{Comparisons of performance and computational cost  among HCMS and various baselines including using a single modality and leveraging multimodal information with early fusion by LSTM networks for 10 and 128 steps respectively.}
  \label{tab:main}
  \resizebox{1.0\linewidth}{!}{
  \begin{tabular}{cc||cc|cc||cc|cc}
    \toprule
    \multirow{ 2}{*}{Method} & \multirow{ 2}{*}{Modal} &\multicolumn{4}{c||}{\fcvid} & \multicolumn{4}{c}{\anet}\\
    \cmidrule(lr){3-6}\cmidrule(lr){7-10}
     &&10 views&GFLOPs&128 views&GFLOPs&10 views&GFLOPs&128 views&GFLOPs\\
    \midrule 
    \multirow{ 7}{*}{LSTM} &A &22.5\%  &0.8 & 27.6\% &10.3 &15.9\%&0.8 & 17.3\%&10.3  \\
    & I &81.4\% &10.1&82.6\%&129.7&76.1\% &10.1& 77.6\% &129.7\\
    & M &85.3\% &657.5&85.6\% &8416.4&86.1\%&657.5& 85.7\% &8416.4 \\ 
    & A + I &84.3\%&10.8&85.5\% &138.6&77.5\%&10.8& 78.8\%&138.6\\
    & A + M &87.6\%&658.0&88.0\% &8421.9 &86.5\%&658.0& 87.0\%&8421.9 \\
    & I + M &86.8\%&667.2&87.0\% &8539.9&87.0\% &667.2& 87.0\% &8539.9\\ 
    & A + I + M &88.4\%&667.9&88.4\%&8549.4& \textbf{87.5\%}&667.9& 87.3\% &8549.4 \\
    \midrule
    \system &A + I + M&\multicolumn{2}{c}{mAP: \textbf{88.8\%}} & \multicolumn{2}{c||}{GFLOPs: 377.4} & \multicolumn{2}{c}{mAP: \textbf{87.5\%}}&\multicolumn{2}{c}{GFLOPs: 576.7}\\
    \bottomrule
  \end{tabular}
  }
\end{table*}

\begin{figure}[b!]
\centering
\subfigure[\fcvid]{
\includegraphics[width=0.45\linewidth]{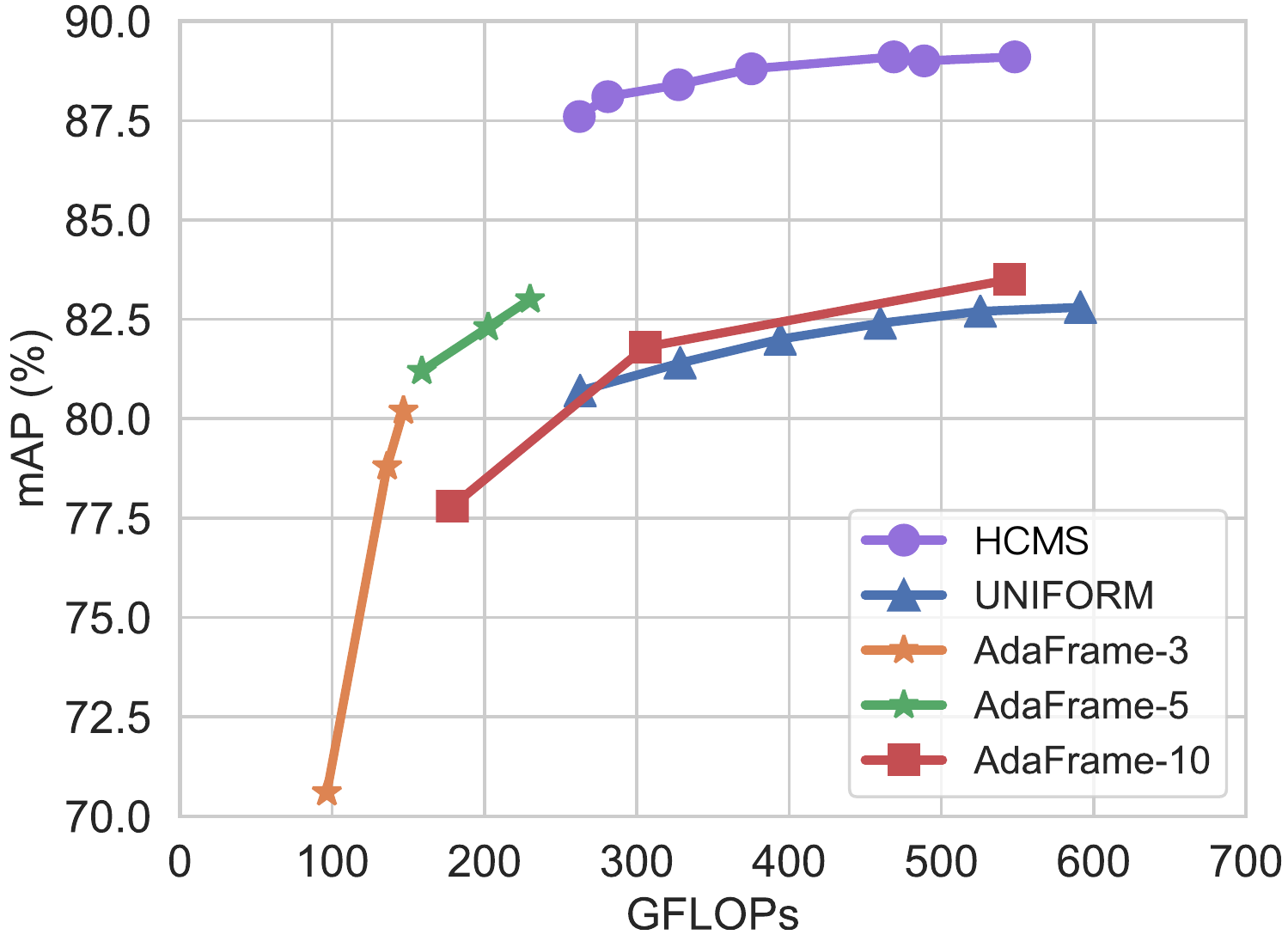}
}
\subfigure[\anet]{
\includegraphics[width=0.45\linewidth]{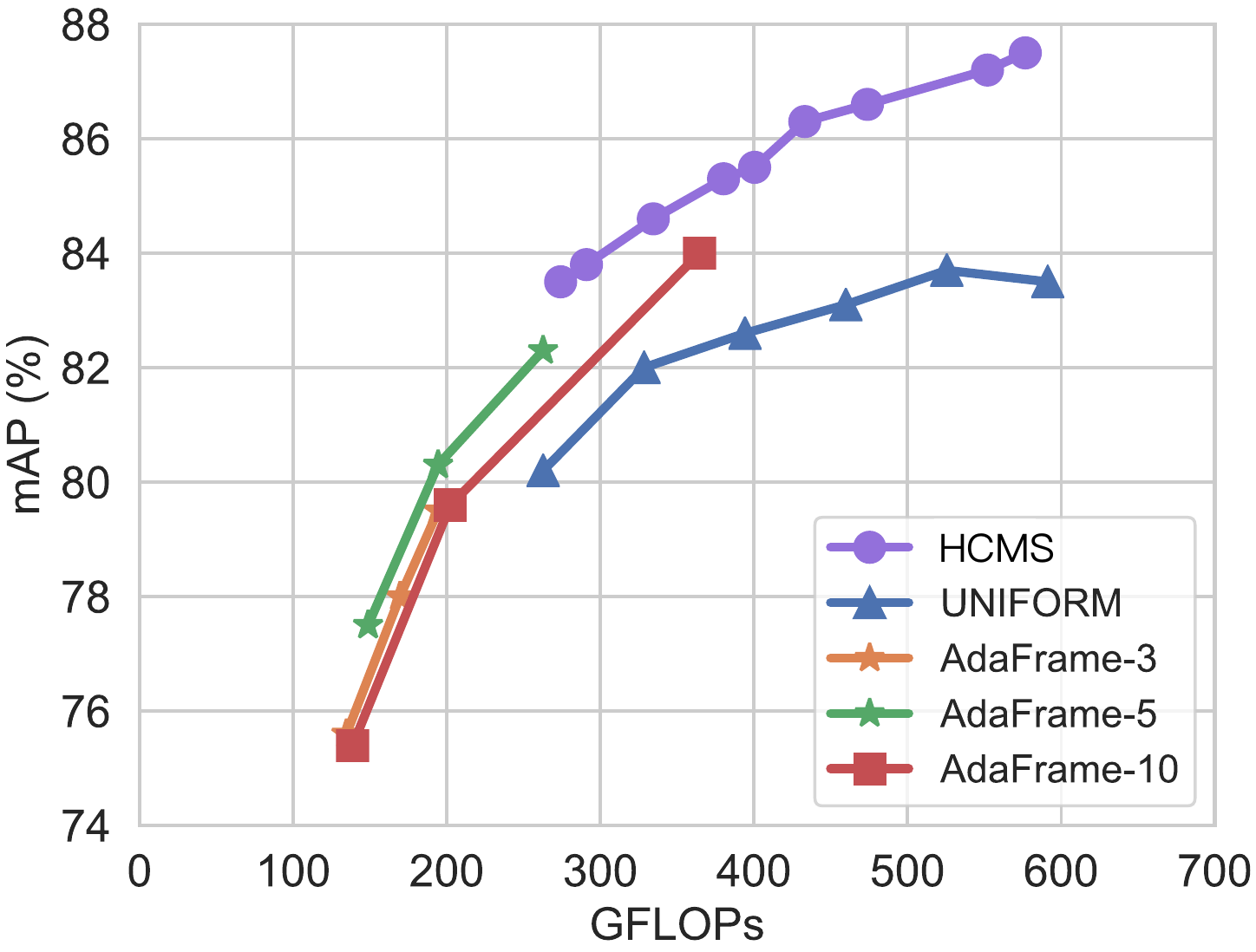}
}
\caption{Computational cost \vs mAP on \fcvid
and \anet.}
\label{fig:cost}
\end{figure}

\subsection{Main Results}

\subsubsection{Effectiveness of modality fusion}
To better understand the the effectiveness of our approach, we first report the recognition performance when using a single modality. To leverage multimodal information, we also combine features with early fusion, which concatenates different modalities as inputs to the LSTM model, as indicated by ``+''. In this set of experiments, we train LSTM networks for 10 and 128 steps, respectively. Table~\ref{tab:main} summarizes the results. We can see from the table that the performance of the audio modality is significantly lower on both datasets (less than 30\% mAP), compared to visual clues captured by appearance and motion networks. However we observe that when adding it on top of appearance or/and motion features, the results consistently improve by $2-3\%$. This confirms that although audio signal itself is not as discriminative as visual cue, it provides useful and complementary context information on top of visual cue for video recognition. 
As demonstrated in Table~\ref{tab:main}, the combination of two modalities steadily improves single-modal baselines, and using all three clues offers the best performance. 

Instead, exploring multimodal information at the instance-level, \system obtains even better performance compared to the early fusion of three features ($88.8\%$ vs.\ $88.4\%$ on \fcvid, $87.5\%$ vs.\ $87.5\%$ on \anet) with significantly reduced computational cost. This highlights that learning to select modality conditioned on inputs samples not only improves the recognition performance but also saves computation.

\begin{figure}[t!]
\centering

\begin{minipage}[t]{.48\textwidth}
  \centering
    \setlength{\tabcolsep}{2pt}
    \renewcommand\arraystretch{1.0}
    \centering
        \tabcaption{Extended application of HCMS in computationally limited online prediction scenarios, where the computational budget for predicting each video is strictly defined. We test HCMS under different computational constraints on \fcvid and \anet.}
         \vspace{0.1in}
    \resizebox{1.0\textwidth}{!}{
      \begin{tabular}{ccc||ccc}
    \toprule
    \multirow{ 2}{*}{GFLOPs} & \multicolumn{2}{c||}{\fcvid} & \multirow{ 2}{*}{GFLOPs} & \multicolumn{2}{c}{\anet} \\
     \cmidrule(lr){2-3}\cmidrule(lr){5-6} 
     & mAP & Acc&& mAP & Acc \\
    \midrule 
    \textless 80 &80.4\% &80.6\%&        \textless80& 64.9\% & 69.2\%   \\
    \textless 120 & 82.4\% & 81.6\%&      \textless140& 75.3\% & 71.9\%  \\
    \textless 160 &86.2\% &84.4\% &       \textless200& 80.9\% & 75.1\%  \\
    \textless 200 & 86.5\% &84.6\% &      \textless260& 83.0\% & 76.9\%  \\
    \textless 240 &87.5\% &85.6\% &       \textless320& 84.2\% & 78.5\%  \\
    \textless 280 &87.6\% &85.8\% &       \textless380& 84.8\% & 79.4\%  \\
    \textless 320 &88.0\% &86.3\% &       \textless440& 85.2\% & 79.7\%  \\
    \textless 360 &88.1\% &86.5\% &       \textless500& 85.5\% & 80.2\%  \\
    \textless 400 &88.2\% &86.7\% &       \textless560& 85.9\% & 80.7\%  \\
    \textless 440 &88.3\% &86.8\% &       \textless620& 86.3\% & 81.1\%  \\ 
    \midrule
     \textless $\infty$ & 88.8\% &87.3\%& \textless $\infty$ & 87.5\%&82.4\% \\
    \bottomrule
  \end{tabular}
      }
    \label{tab:online}
    
    \setlength{\tabcolsep}{1.4pt}
\end{minipage}
\quad
\begin{minipage}[t]{.48\textwidth}
    \setlength{\tabcolsep}{4pt}
    \renewcommand\arraystretch{1.0}
    \centering
    \tabcaption{Comparisons among HCMS and other state-of-the-art methods on \fcvid and \anet.\\}
    \vspace{0.1in}
    \resizebox{1.0\textwidth}{!}{
        \begin{tabular}{c|cccc}
         
        \toprule
        Method  & \fcvid & \anet\\
        \midrule
        IDT~\cite{wang2013action} &-& 68.7\% \\ 
        C3D~\cite{C3D}            &-& 67.7\% \\ 
        P3D~\cite{qiu2016deep}  &-& 78.9\%   \\ 
        RRA~\cite{zhu2018fine}  &-&83.4\% \\ 
        
        MARL~\cite{wu2019multi} &-& 83.8\% \\ 
        
        GSFMN~\cite{zhao2019visual} & 76.9\% &- \\
        Pivot CorrNN~\cite{kang2018pivot}& 77.6\%&- \\
        
        AdaFrame~\cite{wu2019adaframe}& 83.0\% & 84.2\% \\
        
        TSN~\cite{wang2018temporal}& -& 76.6\% \\
        KeylessAttention~\cite{long2018multimodal}& -&78.5\% \\
        FV-VAE~\cite{qiu2017deep}& -&84.1\% \\
        IMGAUD2VID~\cite{gao2020listen}&- & 84.2\% \\
        CKMN~\cite{qi2020towards}& 81.7\% & 85.6\%\\
        \midrule
        \system &\textbf{88.8\%} & \textbf{87.5\%}  \\
        \bottomrule
      \end{tabular}
    }
    \label{tab:sota}
\end{minipage}

\end{figure}

\subsubsection{Performance under Different Computational Budgets} \system can accommodate different computational budgets at both the training stage and the testing stage.

\vspace{1mm}
\textbf{Training stage}. As $\gamma1$ and $\gamma2$ in Eqn.~\ref{eqn:loss_usage} control the targeted computational budget, a series of models under different computational costs could thus be readily produced by varying the values of $\gamma1$ and $\gamma2$. As demonstrated in Fig.~\ref{fig:cost}, \system can cover a wide range of trade-offs between recognition performance and efficiency, and consistently outperforms the UNIFORM baseline under different computational costs by clear margins. To clarify,the computational budgets of baseline methods are controlled by changing the number of sampled views per video. Additionally, we observe that \system also outperforms AdaFrame~\cite{wu2019adaframe}. This confirms that our method can flexibly provide a group of models for video recognition with different computational budgets. 

\vspace{1mm}
\textbf{Testing stage}. During testing, a computational budget can be specified and \system emits a prediction once it consumes all the pre-defined budget. This is particularly useful when deploying a trained model in online settings. The results are summarized in Table~\ref{tab:online}. We can see that even with limited computational budgets (less than 80 GFLOPs), \system achieves a decent classification performance on both datasets. This suggests that \system can be deployed in an online setting.

\begin{figure}[t!]
\centering
\includegraphics[width=0.8\linewidth]{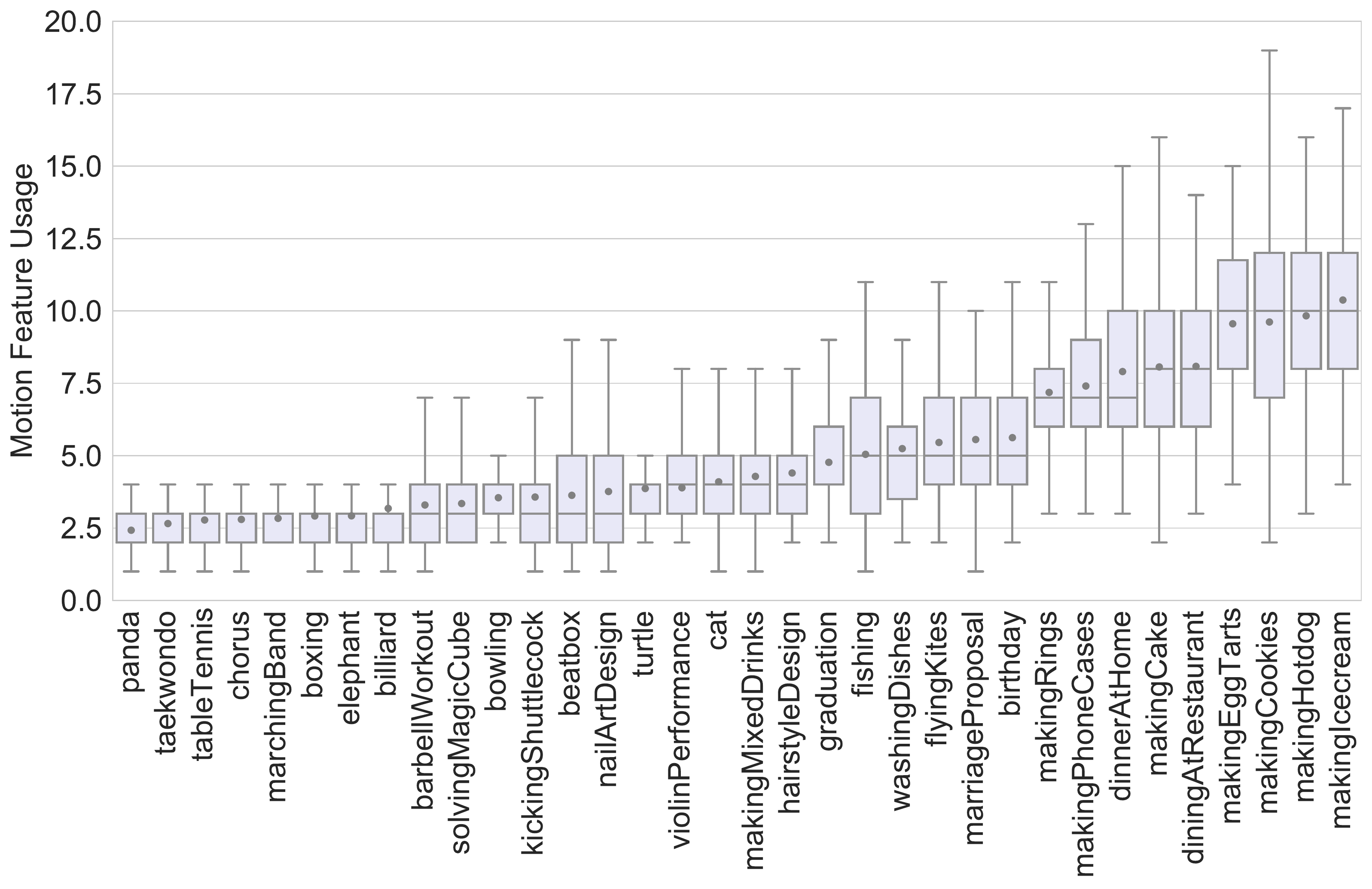}
\caption{The distribution of motion feature usage for sampled classes on FCVID. In addition to quartiles and medians, mean usage, denoted as gray dots, is also presented.}
\label{fig:distribution_motion}
\end{figure}

\begin{figure*}[t!]
\centering
{
\includegraphics[width=0.49\linewidth]{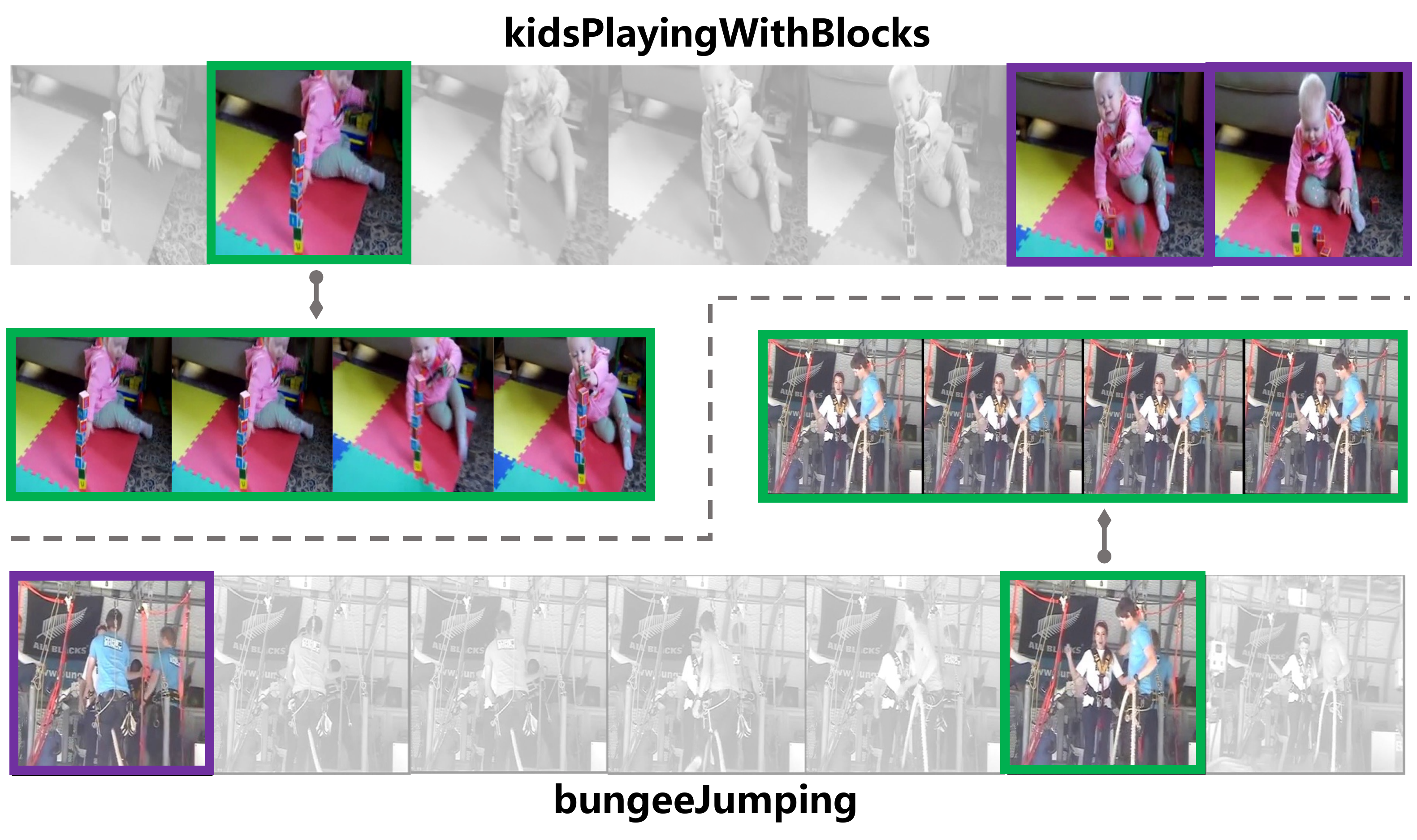}
}
{
\includegraphics[width=0.49\linewidth]{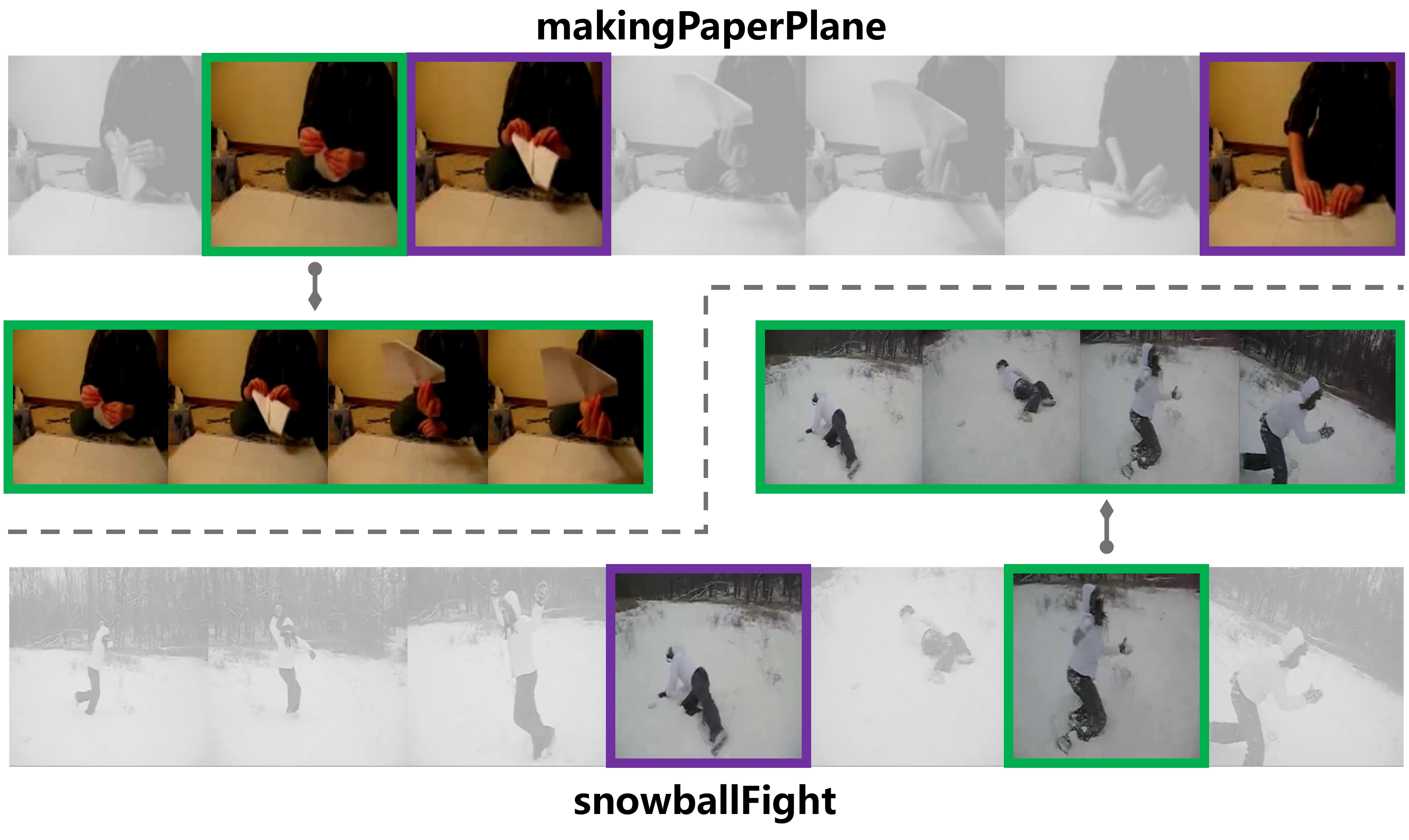}
}

\caption{Frames/clips selected by \system for sampled videos from \fcvid. Here, green borders indicate the use of the motion modality whereas purple borders represent the use of the appearance modality.} 
\label{fig:qualitative}
\end{figure*}

\subsubsection{Comparison with the state-of-the-art}  We now compare \system against various approaches for video recognition. In particular, we compare it with several adaptive computation methods for efficient video recognition like AdaFrame~\cite{wu2019adaframe}, MARL~\cite{wu2019multi} and IMGAUD2VID~\cite{gao2020listen}. Specifically, AdaFrame and MARL learn to produce dynamic frame/clip usage policies such that fewer time steps are sampled from videos for efficient inference, and IMGAUD2VID further proposes to use audio information to efficiently preview videos and quickly reject non-informative clips instead of using them all. Standard methods that forgo adaptive computation such as C3D~\cite{C3D}, TSN~\cite{wang2018temporal}, RRA~\cite{zhu2018fine}, CKMN~\cite{qi2020towards} are also compared. Results are summarized in Table~\ref{tab:sota}. We can see that \system obtains a state-of-the-art performance of $88.8\%$ and $87.4\%$ mAP on \fcvid and \anet respectively, and outperforms all the other methods listed. It is worth pointing out that our method is orthogonal to adaptive methods on selecting salient frames/clips, input resolutions and salient spatial regions~\cite{wu2019adaframe,wu2019multi,korbar2019scsampler,meng2020ar,uzkent2020learning}, and thus could be complementary to them.

\begin{figure}[t!]
\centering
\subfigure[Appearance Feature Selection]{
\includegraphics[width=0.48\linewidth]{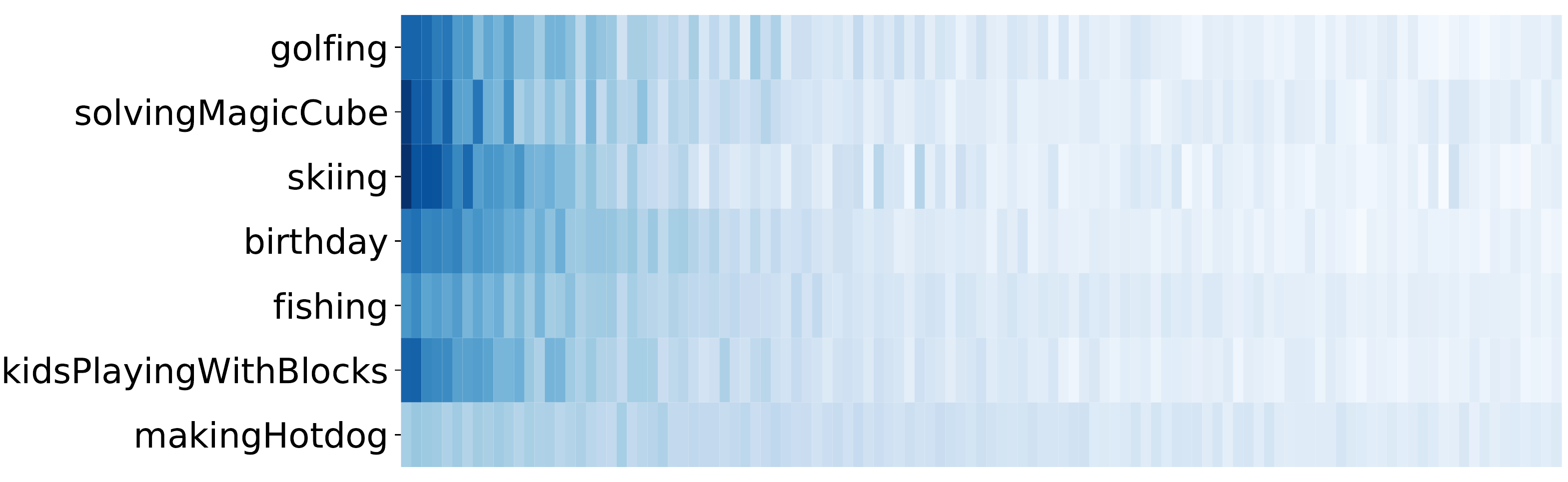}
}
\subfigure[Motion Feature Selection]{
\includegraphics[width=0.48\linewidth]{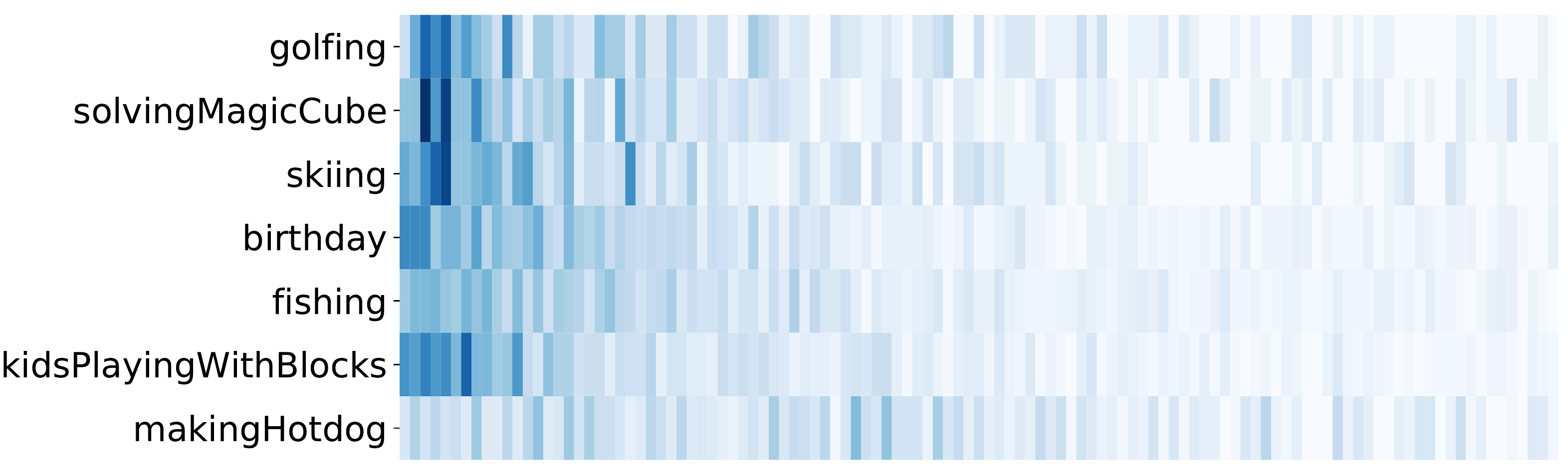}
}
\caption{Time steps (left $\rightarrow$ right) selected by \system to use (a) appearance feature and (b) motion feature for different action categories. Darker color denotes more frequent usage of fine feature at the corresponding time step. Best viewed in color.}
\label{fig:temporal}
\end{figure}

\begin{table*}[t!]
\centering
  {{
  \caption{Model using the audio information only. Some categories are quite sensitive to audio and can have high AP scores compared to the overall AP score.}
  \label{tab:audiosensitive}
  \resizebox{0.9\linewidth}{!}{
  \begin{tabular}{cc||cccccccc}
    \toprule
    \multirow{ 2}{*}{Method} & \multirow{ 2}{*}{Metric} & \multirow{ 2}{*}{Overall$^{*}$} & Playing & Playing & Playing & Playing & Drum & Playing  & Playing \\ 
    &&  & bagpipes &  accordion & saxophone & flauta & corps  & congas & violin  \\
    \cmidrule(lr){1-2}\cmidrule(lr){3-3}\cmidrule(lr){4-10}
    Audio LSTM & AP & 17.3\% & 96.1\% & 88.0\% & 87.4\% & 81.5\% & 79.4\% & 76.0\% & 72.1\%\\
    \bottomrule
  \end{tabular}}
  }
  }
\end{table*}

\begin{table*}[t!]
\centering
  {{
  \caption{HCMS performance on audio-sensitive categories.}
    \label{tab:HMSaudiosensitive}
  \resizebox{0.9\linewidth}{!}{
  \begin{tabular}{cc||cccccccc}
    \toprule
    \multirow{ 2}{*}{Method} & \multirow{ 2}{*}{Metric} & \multirow{ 2}{*}{Overall$^{*}$} & Playing & Playing & Playing & Playing & Drum & Playing & Playing  \\ 
    & & & bagpipes &  accordion & saxophone & flauta & corps & congas & violin  \\
    \cmidrule(lr){1-2}\cmidrule(lr){3-3}\cmidrule(lr){4-10}
    \multirow{ 2}{*}{HCMS} & AP & 87.5\% & 93.7\% & 98.2\% & 94.6\% & 91.5\% & 93.8\% & 96.3\% & 83.6\% \\
    & GFlops & 576.7 & 341.1 & 254.0 & 456.4 & 423.3 & 295.9 & 224.1 & 441.5\\
    \bottomrule
  \end{tabular}}
  }
  }
\end{table*}

\subsection{Discussion}

Having demonstrated the effectiveness of HCMS, we now conduct quantitative and qualitative analysis to probe how it helps improve efficiency without sacrificing recognition performance. 

\subsubsection{Learned Usage Strategies}

We first analyze learned usage strategies by examining how much higher-cost modalities are utilized for each category. For this purpose, statistics of learned motion feature usage on \fcvid are displayed in Figure~\ref{fig:distribution_motion}. It can be seen the learned usage strategies have large both inter-class and intra-class variances. For some categories with strong audio or appearance cues like ``Chorus'', ``Panda'' and ``Billiard'', our method allocates quite low usage of motion features. On the contrary, the complex and visually similar categories are utilizing more motion features. For example, the audio and appearance features for ``making egg tarts'', ``making cookies'' and ``making ice cream'' could be very similar as they mostly occur in the same surroundings (\ie ``kitchen''), for which motion information is critical to distinguish the subtle movements.

Visualizations of selected time steps for using finer appearance and motion features, are shown in Figure~\ref{fig:temporal}. They confirm the observations above as well. For instance, motion-intensive classes like ``making hotdog'' tend to use motion features more often across the entire video, yet a few glimpses of visual appearance for scene-centric classes like ``golfing'' are already sufficient.

\subsubsection{Qualitative Analysis} We further visualize the time steps selected by \system to use appearance and motion features for qualitative analysis. As shown in Figure~\ref{fig:qualitative}, frames and clips with unique appearance and motion cues tend to be selected, while similar and still visual patterns are more likely to be skipped. For instance, we see from Fig.~\ref{fig:qualitative}, redundant frames are skipped when the person in white shirt is occluded.

\subsubsection{Assumption Revisit} In the introduction, we mentioned that (1) different video categories require different salient modalities; (2) modality needed differ even for samples within the same category; (3) audio modality is critical for some videos. In Figure~\ref{fig:distribution_motion}, we show the number of modalities selected by HCMS differs across categories while in Figure~\ref{fig:temporal}, the selection pattern of videos belonging to the same category also shows diversity. Here, we verify that visual information is less needed for actions with clear audio cues. We firstly use an audio-only model to find the audio-sensitive categories as shown in Table~\ref{tab:audiosensitive}. Despite the fact that the overall 17.3\% mAP from the audio information is quite low, some categories achieve high AP scores because they are audio-sensitive and may need fewer visual clues to make precise predictions. To verify this assumption, we evaluate HCMS on those categories and find that their corresponding computational cost is much lower than the average as shown in  Table~\ref{tab:HMSaudiosensitive}.

\begin{table*}[t!]
\centering
  \setlength{\tabcolsep}{4pt}
  \renewcommand\arraystretch{1.2}
  \caption{Analysis of the modality order in our designed hierarchical architecture. ``SkipRatio1'' and ``SkipRatio2'' correspond to the testing skipped proportion of the middle and tail modalities in a specific modality order. $A\rightarrow I\rightarrow M^*$ denotes the baseline order.}
  \label{tab:discussgeneral}
  \begin{tabular}{cl||cccc}
    \toprule
    Method & Modality order & ~\  mAP \  ~ & ~GFLOPs~ & SkipRatio1 & skipRatio2 \\
    \midrule
    \multirow{ 4}{*}{\system} & ~ A $\rightarrow$ I $\rightarrow$ M $^{*}$ & 87.5\% & 576.7 & 0.525 & 0.940   \\
    \cmidrule(lr){2-6}
    & ~ I $\rightarrow$ A $\rightarrow$ M & 87.7\% & 587.4 & 0.583 & 0.946 \\
    & ~ A $\rightarrow$ M $\rightarrow$ I & 88.2\% & 2898.7 & 0.658 & 0.942  \\
    & ~ M $\rightarrow$ I $\rightarrow$ A & 88.4\% & 8474.9 & 0.603 & 0.942  \\
    \bottomrule
  \end{tabular}
  
\end{table*}

\subsubsection{Discussion of the generalizability of \system} 
We build HCMS by an interactive hierarchical LSTMs design whose target is to save computational cost without compromising the performance when recognizing videos with multimodal information. Along the line of saving the amount of computational resources, the first important application extension of HCMS is online inference with limited computational budget, as mentioned in previous experiments. Since HCMS guarantees future frames cannot be observed at the current moment, it is in fact well suited for the mobile video streaming scenario with limited computational resources. HCMS offers good predictions before the computational budget is surpassed. Besides, HCMS is not only capable of handling multimodal video tasks, but also has the potential to handle more diverse multimodal streaming tasks. There is no difficulty in applying HCMS to other tasks with rich temporal information. The key challenge is whether HCMS can still learn good selection patterns with different modalities, e.g., using a different modality order. We provide some positive evidence for such a challenge from the perspective of modality order. In HCMS, we set the order as Audio-Image-Motion due to their computational consumption. We intuitively elaborate the rationality of the such combination order and have verified the reasonableness by many quantitative results and visualization examples already. However, we argue HCMS is a flexible and compatible framework that will not be useful only in a particular specified modal order. 
To demonstrate this, we arbitrarily switched the two modality orders based on Audio-Image-Motion and keep the skipping constraint the same in the training stage. Experimental results are shown in Table~\ref{tab:discussgeneral}. Here, $A\rightarrow I\rightarrow M^*$ denotes the baseline order. 
Obviously, using computationally expensive features as inputs to the shallow LSTM branch leads to an increase in computation. In addition, HCMS is able to perform consistently well in different orders, which reflects that HCMS offers good selection patterns under different settings and has the potential to be easily applied to other multimodal streaming tasks.

\subsubsection{Training curves under different selection constraints.} 
\begin{figure}[t!]
\centering
\includegraphics[width=1.0\linewidth]{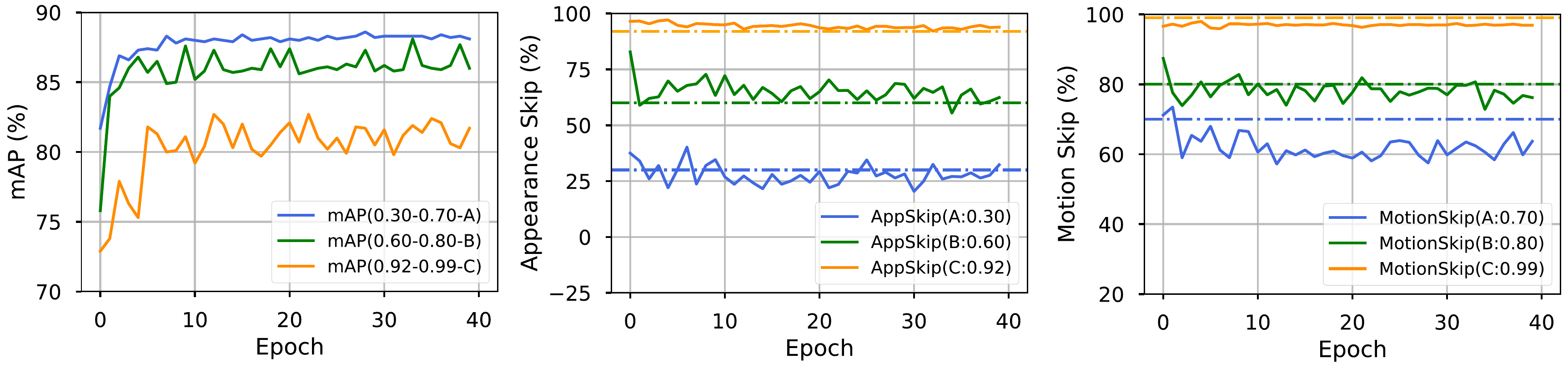}
\caption{Training curves under different selection constrains. \system operates on the low-cost audio modality by default, and at each time step, it learns dynamically whether to use high-cost modalities on a per-input basis. Here, A/B/C represent three different appearance/motion skip ratio groups.
}
\label{fig:traincurve}
\end{figure}

Training phase is affected by the modality selection.  Intuitively, the larger appearance/motion skip ratios produce worse results. To verify this, we set three different appearance/motion skip ratio groups, which are (0.3, 0.7), (0.6, 0.8), and (0.92, 0.99) and visualize the trend of accuracy and the modality skip ratios throughout the training process as shown in Figure~\ref{fig:traincurve}. We can see the line graphs in the second and third panels, where the solid lines indicate the modality skipping frequencies learned by the gating module at different training epochs, and the dashed lines indicate the expected skip ratios. It is found that the gating module learns very quickly about how many modality features should be selected in the early training stage. However, as is shown in the first panel line graph, the mAP score is relatively low in the early stage and gradually increases as the training proceeds, indicating the great importance of selecting key frames for each modality. Overall, the higher skipping ratios we constrain, the worse the model performs as shown in the first panel, in line with our aforementioned intuition.

\section{Conclusion}
In this paper, we presented Hierarchical and Conditional Modality Selection (HCMS), which is a conditional computation framework for multimodal learning on video recognition. In contrast to conventional video recognition approaches that leverage multimodal features for all samples, we learn what modalities to use on a per-input basis. The intuition of \system is to use a low-cost modality by default, and learns whether to compute high-cost features conditioned on each input sample. In particular, \system contains an audio LSTM, an appearance LSTM and a motion LSTM working collaboratively in a hierarchical manner. The audio LSTM operates on audio information to get an overview of video contents. The appearance and the motion LSTM contains gating functions to decide whether to use its corresponding modality. Extensive experiments are conducted on \fcvid and \anet. We demonstrate that \system achieves state-of-the-art performance on both datasets with significantly reduced computational cost. 

\section{Acknowledgement}
This project was supported by NSFC under Grant No. 62102092 and Shanghai Science and Technology Program (Project No. 21JC1400600).
 
\bibliographystyle{ACM-Reference-Format}
\bibliography{main}

\end{document}